\documentclass[arxiv]{clear2026}

\usepackage{algorithm}
\usepackage{algorithmic} 
\usepackage{mathtools} 
\usepackage{booktabs} 
\usepackage{times}
\usepackage{pdfpages}

\usepackage{tikz}
\usepackage{multirow} 
\usepackage{graphicx}

\usepackage{dirtytalk}

\title{Adversarial Causal Tuning for Realistic Time-series Generation}

\clearauthor{%
 \Name{Nikolaos Gkorgkolis$^*$} \Email{gkorgkolis@csd.uoc.gr} \\
 \Name{Nikolaos Kougioulis$^*$} \Email{kougioulis@csd.uoc.gr} \\
 \Name{Ioannis Tsamardinos} \Email{tsamard@csd.uoc.gr} \\
 \addr University of Crete \& FORTH, Heraklion, Greece
 \AND
 \Name{Mingxue Wang} \Email{wangmingxue1@huawei.com} \\
 \Name{Bora Caglayan} \Email{bora.caglayan@huawei.com} \\
 \Name{Andrea Tonon} \Email{andrea.tonon1@huawei-partners.com} \\
 \Name{Dario Simionato} \Email{dario.simionato1@h-partners.com} \\
  \addr Huawei Ireland Research Centre, Dublin, Ireland
}

\begin{document}

\def\thefootnote{*}\footnotetext{Authors contributed equally to this work.}\def\thefootnote{\arabic{footnote}}
\renewcommand{\thefootnote}{\arabic{footnote}}

\maketitle

\begin{abstract}
We address the problem of generating simulated, yet realistic, time-series data from a causal model with the same observational and interventional distributions as a given real dataset (probabilistic causal digital twin). While non-causal models (e.g., GANs) also strive to simulate realistic data, causal models are fundamentally more powerful, able to simulate the effect of interventions (what-if scenarios), optimize decisions, perform root-cause analysis, and counterfactual causal reasoning. We introduce the Adversarial Causal Tuning (ACT) methodology, which outputs the optimal causal model that fits the data, along with a quantification of the goodness-of-fit. The returned causal model can then be employed to simulate new data or to perform other causal reasoning tasks. ACT adopts ideas from Generative Adversarial Network training and AutoML to search for optimal causal pipelines and discriminators that detect deviations between the distributions of real and simulated data. It also adapts a permutation testing procedure from established causal tuning methods to penalize models for complexity. Through extensive experiments on real, semi-synthetic, and synthetic datasets, we show that (a) employing multiple optimized discriminators is paramount for selecting the optimal causal models and quantifying goodness-of-fit, (b) ACT selects the optimal causal model in synthetic datasets while avoiding overfitting, generating data indistinguishable from the true data distribution (c) all state-of-the-art generative and causal simulation methods, exhibit room for improvement in reproducing real data distributions; generating realistic temporal data is still an open research challenge.

\end{abstract}

\begin{keywords}
Causal Discovery, Causal model selection, Generative Models, Time-series
\end{keywords}

\section{Introduction \label{sec:introduction}}

Causal models encode the generative mechanisms underlying the data. Unlike predictive or generative models, they support the simulation of interventions to explore what-if scenarios, decision optimization, root-cause analysis, and counterfactual reasoning \citep{pearl2009causality}. Such models are rarely fully specified based on prior knowledge alone. Hence, causal discovery algorithms have been developed to induce them from observational data, or a limited number of interventional data \citep{spirtes2001causation, runge2018causal, runge2019inferring}. Numerous such algorithms have been proposed in the literature \citep{ZANGA2022101}. However, no universally optimal algorithm exists: the efficacy of an algorithm depends on the input dataset. Therefore, one should search the space of causal modeling pipelines to tune algorithmic choices and their hyperparameters to identify the best-fitting causal model, giving rise to the \textbf{causal tuning} problem. The latter is conceptually equivalent to tuning predictive models in automated machine learning (AutoML) systems. However, causal tuning is an \textit{unsupervised task} which requires the development of special methodologies.

In this work, we present a general principle for causal tuning called \textbf{Adversarial Causal Tuning} or \textbf{ACT}. Given as input a multiple time-series dataset, it returns an optimal temporal causal model and a measure of goodness-of-fit. Our main motivation for ACT in this paper is \textit{realistic data generation}. Specifically, it is to identify a causal model that generates data indistinguishable from the real ones. Crucially, it should also be able to simulate what-if scenarios where one intervenes to set the value of a variable and produce data from the interventional distribution \(P(X|\mathrm{do}(C=c))\). Such \textit{a causal model can thus serve as a probabilistic Causal Digital Twin and a foundation for privacy-preserving data sharing} \citep{homaei2025causal}. Notice that non-causal generative models, such as conditional GANs \citep{mirza2014conditional}, can generate data from the conditional distribution \(P(X|C=c)\) which is not always equal to \(P(X|\mathrm{do}(C=c))\). 

ACT also aims to solve another fundamental problem in causal discovery, namely, realistic benchmark creation for causal algorithms. Specifically, to benchmark a causal algorithm, one needs to have available a dataset as well as its corresponding ground-truth, data-generating causal model. Existing benchmarks are almost exclusively comprised of \textbf{synthetic causal models}. These are produced by randomly sampling the causal graph from a graph distribution (e.g., Erd\H{o}s–Rényi or Barabási–Albert distributions \citep{brouillard2020differentiable}). The causal functional dependencies are sampled from small hand-crafted families (e.g., linear Gaussian or simple nonlinear additive noise models), while the noise distributions are typically parametric. \textit{It is highly questionable whether the community corpus of benchmark causal models resembles real causal models encountered in practice}. ACT can be employed to fit real datasets; whenever an acceptable goodness-of-fit has been achieved by the winning causal model, it can then be added to a benchmark corpus of realistic causal models. Finally, ACT can serve as the heart of automated causal discovery platforms (\textbf{AutoCD}) \citep{biza2024towards} that try to automate the induction of causal models and causal inference tasks. 

ACT frames causal model selection as a minimax problem: a search in the space of causal pipelines (\textit{causal configurations}) tries to identify a causal model that \textit{minimizes} the ability of discriminators to distinguish real from simulated data; simultaneously, a search in the space of predictive pipelines (\textit{discrimination configurations}) tries to identify the discriminator that \textit{maximizes} the identification of the source of a sample: real or simulated. If the best discriminator performs no better than random guessing for a causal model, the latter achieves goodness-of-fit. 
ACT also includes a sparsity penalty that prevents overfitting, based on permutation testing \citep{biza2022oct}. ACT ensures distributional fit rather than causal fit; the true causal graph may be different, but it should at least be a Markov equivalent one \citep{spirtes2001causation}. Causal fit can only be validated with experiments. ACT is comparable to a model-agnostic and discrete version of the GAN minimax procedure. The searches in the causal and discrimination configuration spaces can employ AutoML techniques \citep{hutter2019automated} such as black-box optimization and meta-level learning. The current implementation of ACT addresses the tuning of Temporal Structural Causal Models, assuming causal sufficiency, no contemporaneous causal effects, and additive noise. 

Our motivational experiments show that one should optimize over discriminators, or risk underfitting. Using a single discriminator is common in the literature and has led prior works to \textit{falsely} conclude that their methodologies can fit models that generate realistic data \citep{cheng2024causaltime}. Our experiments show that this conclusion could be an artifact of the lack of optimization. 

Our contributions (a) include the formalization of ACT and an open-source implementation {\texttt{(url withheld to maintain anonymity)}. (b) We show that penalizing for sparsity is essential for preventing overfitting of the selected model. (c) We demonstrate that ACT outperforms state-of-the-art methodologies in fitting datasets from synthetic causal models. In addition, (d) ACT can most often successfully generate realistic data under interventions. Finally, (e) we show that on real datasets, all state-of-the-art methods fail to successfully fit a generative model. Realistic time-series simulation remains an open challenge, motivating the need for improved causal discovery algorithms. ACT can be used for generating realistic data, for causal model selection within automated causal discovery pipelines, and for creating realistic causal models for benchmarking algorithms.

\section{Preliminaries \& Problem Formulation \label{subsec:preliminaries}}

\begin{figure}[t!]
\centering

\begin{minipage}[b]{0.4\textwidth}
\centering

\includegraphics[width=\textwidth]{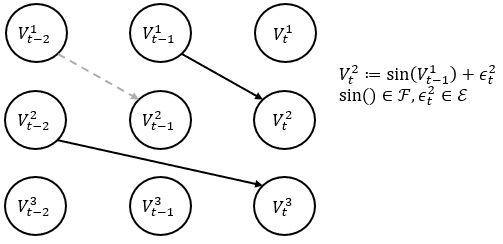}
{(a) Lagged Causal Graph}
\end{minipage}
\hfill
\begin{minipage}[b]{0.4\textwidth}
\centering
\includegraphics[width=\textwidth]{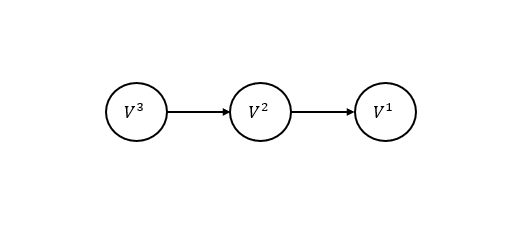}

{(b) Summary Causal Graph}
\end{minipage}
\caption{(a) The lagged graph of a TSCM with max lag \(\ell_\text{max}=2\) where \(V^1\) causes \(V^2\) with lag 1 and \(V^2\) causes \(V^3\) with lag 2. The functional dependency of \(V^2_t\) on its direct causes (parents) is shown. (b) The corresponding summary causal graph (as in \citep{cheng2024causaltime}) discards time lag information, representing only the aggregated causal influences.}
\label{fig:temporal-graphs}
\end{figure}

\paragraph{Temporal structural causal models (TSCM)} We consider a discrete multivariate process of a vector of variables \(V \in \mathbb{R}^N\) over time points \(1, \ldots, T\). \(V^j_t\) denotes the value of variable \(V^j \in \mathbf{V}\) at time \(t\), while \(\mathbf{V}^j_{<t}\) denotes the past values of \(V^j\), i.e., the set \(\{V^j_1, \ldots, ... V^j_{t-1}\}\). We assume direct causal relations are given by structural equations of the form: \( V^j_t := f^j(\mathrm{Pa}(V^j_t))+ \epsilon^j_t \), where \(\mathrm{Pa}(V^j_t) \subseteq \cup_j \{ V^j_{<t}\}\) and is called the \textit{parent set} of \(V^j\). Thus, \(V^j_t\) is causally determined by a set of past values of \(\mathbf{V}\), but not contemporaneous values of \(\mathbf{V}\). The noise terms \(\epsilon^j_t\) are independent of each other and all variables, hence we assume no latent confounding factors (causal sufficiency). We assume that the causal functional relationships \(f^j\) and the distribution of the noise terms do not depend on time. The causal equations can be qualitatively represented with a graph \(G\) where the vertices correspond to each \(V^j_t\) and there is an edge \(V^j_{t'} \rightarrow V^k_t\), whenever \(V^j_{t'} \in \mathrm{Pa}(V^k_t)\). We assume the graph structure does not depend on absolute time, hence we consider only the portion of the graph with variables \(V_{t-\ell_\text{max}}, \ldots, V_{t}\), where \(\ell_\text{max}\) is the maximum lag for which an edge \(V^j_{t-\ell_\text{max}} \rightarrow V^k_t\) exists. We refer to this subgraph as the \textbf{Lagged Causal Graph} of the process. The lagged graph captures the dynamics of the process. A \textbf{TSCM} is a tuple \(\langle G, F, E, C \rangle\), where \(G\) is a lagged causal graph, \(\mathcal{F}=\{f^i, i=1, \ldots, N\}\) the set of functional dependencies, \(\mathcal{E}\) the set of noise distributions, and \(C\) the probability distribution of the initial conditions \(V_1, \ldots , V_{\ell_{\text{max}+1}}\). The \textbf{Summary Causal Graph} is defined as a graph containing a single vertex for every variable and an edge \(V^j \rightarrow V^k\) if \(V^j_{t-\ell} \rightarrow V^k_t\) for some lag \(\ell\). Given that we are mostly interested in capturing the dynamics of the process, we will drop from the model the initial distribution \(C\) and represent TSCMs with the tuple \(\langle G, F, E \rangle\). 

\paragraph{Data generation (simulation) from a TSCM.}
Given a TSCM data can be generated by \textit{ancestral sampling} \citep{murphy2023probabilistic}: parent values are retrieved from previous lags, noise terms are drawn, and structural equations are evaluated at each time step. Initial values \(V^j_t\) for \(t\leq \ell_\text{max}\) are sampled from the distribution \(\epsilon^j\in\mathcal{E}\). Initial warm-up steps are discarded to reduce any noisy dependence on initial conditions.  

\paragraph{Problem statement.} Let \(\{\mathbf{x}_t\}_{t=1}^{T}\), \(\mathbf{x}_t\in\mathbb{R}^N\), denote an observed, real multivariate time-series of \(N\) quantities observed over \(T\) timesteps. \textbf{The objective is to learn a TSCM that generates simulated data whose distribution is indistinguishable from the real data}. Achieving this requires estimating a lagged causal graph, functional causal (structural) relationships, and noise distributions.

\section{Related Work on Realistic Temporal Data Generation\label{sec:related-work}}

\begin{figure}
    \centering
    \begin{minipage}[t]{0.31\textwidth}
        \centering
        \includegraphics[width=\textwidth]{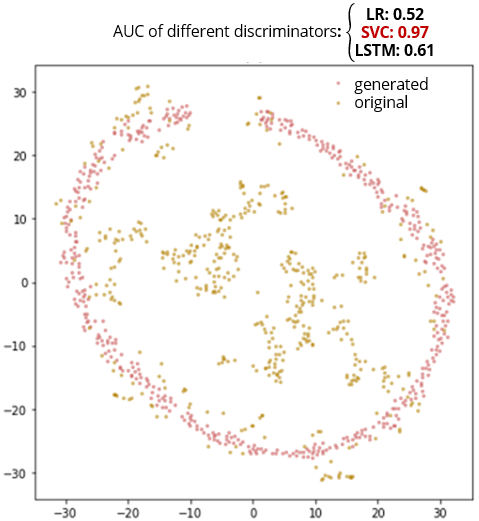}
    \end{minipage}%
    \hfill
    \begin{minipage}[t]{0.31\textwidth}
        \centering
        \includegraphics[width=\textwidth]{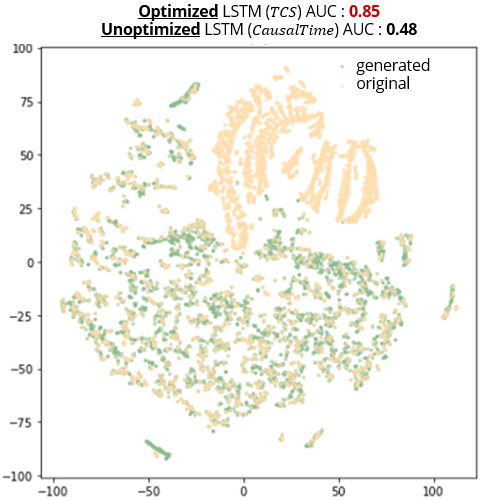}
    \end{minipage}
    \begin{minipage}[t]{0.31\textwidth}
        \centering
        \includegraphics[width=\textwidth]{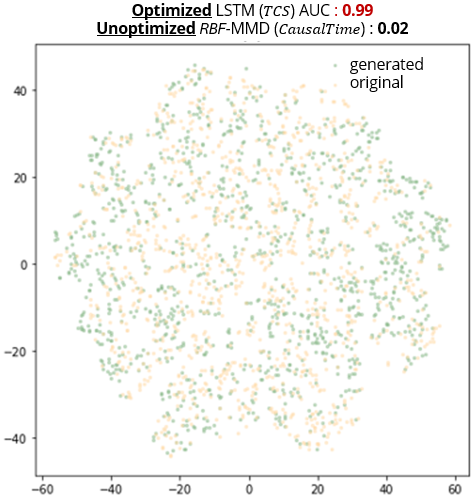}
    \end{minipage}%
    \caption{t-SNE \citep{van2008visualizing} 2D projections of original vs. simulated data for a synthetic, real, and semi-synthetic dataset. (Left) The data distributions are visually different, but one needs to optimize over several discrimination algorithms to identify the one that achieves high AUC. (Middle) The same algorithm (LSTM) is able to distinguish the data distributions when its hyperparameters are tuned properly. (Right) t-SNE is not able to visually depict the differences between the data distributions, but an optimized LSTM achieves discrimination \(\mathrm{AUC}=0.99\). An unoptimized kernel MMD is not able to detect distributional differences. Prior works that rely on unoptimized discriminators and metrics may overclaim successful fitting of real datasets with generative models.}
    \label{fig:t-sne}
\end{figure}

We focus on prior works, causal or non-causal, that, given an input real dataset, generate simulated data whose distributions should be indistinguishable from the real dataset. Regarding non-causal methods, the work of \citet{zhang2022sequential} concerns the CPAR (Conditional Probabilistic Auto-regressive) model for generating multivariate time-series samples, while TimeVAE \citep{desai2021timevae} follows a variational autoencoder \citep{murphy2023probabilistic} approach. Both these methods are non-causal and hence cannot simulate interventional data. CausalTime \citep{cheng2024causaltime} fits a residual-based neural network to model a VAR process and then extracts a \textit{hypothetical causal graph} (using the authors' terminology) using feature-importance methods such as DeepSHAP \citep{lundberg2017unified}. However, the resulting graph is a \textbf{summary graph} without lag information (Figure \ref{fig:temporal-graphs}) and does not correspond to a temporal causal model. As emphasized in their work, \textit{feature-importance scores do not constitute causal discovery, and the generative mechanism learned by their network does not approximate a TSCM}. As the method does not model an explicit TSCM, interventional simulations or counterfactual reasoning are not supported. 

While our focus is the generation of realistic time series data from a causal model, ACT is a general methodology for causal model selection. Prior work on this problem is the Out-of-sample Causal Tuning (OCT) algorithm \citep{biza2022oct}. However, OCT does not focus on the quality of the data generation and is implemented only for atemporal (cross-sectional, i.i.d.) data. Significant adaptations are required to enable the comparison of OCT and ACT as causal tuning methods for temporal data which are left for future work.

\section{Motivation for optimizing over discriminators \label{sec:motive}}

Prior works use different methodologies to assess the successful fitting of a model and its ability to generate realistic data. These methodologies typically rely on using a single discrepancy measure (e.g., MMD) \citep{gretton2012kernel} or discriminator (e.g. SVM, LSTM, LR) \citep{lopez2017revisiting}. However, different discriminators vary dramatically in their statistical power depending on the dataset and hyperparameter configuration. As a result, poorly tuned discrimination can incorrectly suggest that simulated and real data come from the same distribution, even when their discrepancy is visually or statistically evident. For example, CausalTime uses a single, fixed discriminator, leading the authors to believe successful fitting in several real datasets; our experiments in Section \ref{sec:experiments} below disprove these claims. Figure \ref{fig:t-sne} illustrates this issue across a synthetic (based on trigonometric functions (left)), real (AirQualityUCI \citep{ye2024mvts} (middle)), and semi-synthetic dataset (\(f\)MRI from \cite{smith2011network} (right)) (see Section \ref{sec:experiments} for a definition). Generative models were fit using CausalTime on each corresponding dataset. This demonstrates the need to optimize over discrimination algorithms as well as hyperparameters. In addition, visualizations cannot always be trusted to spot distributional differences. 

\section{Adversarial Causal Tuning \label{subsec:act}}

\begin{figure*}
    \centering
    \includegraphics[width=\linewidth]{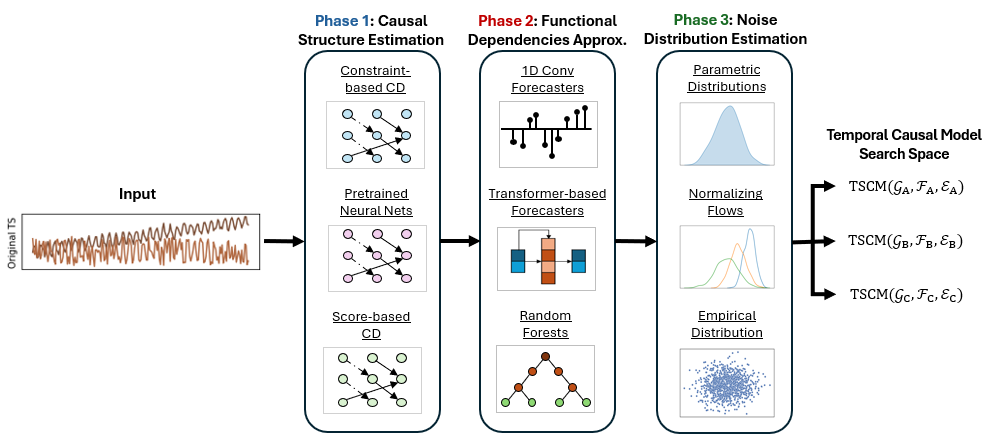}
    \caption{ACT's production of TSCM space: Given a multivariate time-series sample of real data, ACT induces lagged causal graphs (\textit{Phase 1}), functional dependencies (\textit{Phase 2}), and noise distributions (\textit{Phase 3}) using all combinations of available algorithms and their hyperparameters. Each combination of Phases 1–3 produces a temporal causal model.}
    \label{fig:overall_architecture_subs}
\end{figure*}

Adversarial Causal Tuning or ACT seeks to \textbf{identify an optimal TSCM that fits the input dataset} and estimate its goodness-of-fit. The returned TSCM can be employed to generate simulated data under observational and/or interventional conditions. Informally, we define ACT as the following selection principle: 
\begin{quote}
    \textbf{ACT principle}: Given an input dataset \(T\), select as best-performing causal model, the one that, despite our best discrimination efforts, generates data as indistinguishable as possible from the real dataset. Select the simplest causal model whose discrimination performance is statistically equivalent to the best-performing one. 
\end{quote}
We now discuss how to implement this principle for the specific context of TSCMs and multivariate time series data. First, we focus on \textit{measuring and estimating discrimination} between two multivariate time series datasets (empirical distributions) \(T\) and \(T'\) on the same variables. As our metric of choice, we use the Area Under the Receiver's Characteristic Curve or \textbf{AUC} achieved by a discriminator \(D\) when presented with the binary classification problem of distinguishing the source \(T\) or \(T'\) of a sample, denoted as \(d_C(T,T')\). This is equivalent to a Classifier 2-Sample Test (C2ST) \citep{lopez2017revisiting}. To account for the temporal information of data on a generic classifier, a time series dataset is converted to time-lagged format. Specifically, a sample \([V_{t-k}, \ldots, V_t]\) is created for every \(t=\ell_\text{max}+1,\ldots, T\) from the previous \(k\) time steps. For time-series-specific classifiers like LSTM, temporal information is determined by the sequence length parameter, as in CausalTime \citep{cheng2024causaltime}. To estimate \(d_C(T,T')\) we hold out 25\% of the data as a test set and train the discriminator on the remaining 75\%. 

Let us now define the TSCM space \(\mathcal{C}\) as the set of TSCMs under consideration. This set can be produced by pipelines of algorithms and their hyperparameters that accept as input a multivariate timeseries \(T\) and output a TSCM. Any such a pipeline should include algorithms for learning the causal structure, the functional dependencies, and the noise distribution, but could also include steps for data transformation or imputation of missing values. We also define \(\mathcal{D}\) as the space of all discrimination models that can be produced from an input dataset. \(\mathcal{D}\) can be produced by any pipeline that accepts time series data and produces a binary classifier. We denote with \(S_C\) a simulated dataset that can be produced by TSCM \(C\). The \textbf{discrimination performance} of a TSCM \(C\) is defined as \(d_C = \max_{D\in\mathcal{D}} d_D(T,S_C)\). We can then define the \textbf{best-performing} TSCM \(C^*\) as:
\begin{equation}
\label{eq:best-performing}
    C_* = \arg_C\min_{C\in\mathcal{C}} d_C = \arg_C\min_{C\in\mathcal{C}}\max_{D\in\mathcal{D}} d_D(T,S_C)
\end{equation} 
This minimax procedure is conceptually equivalent to the search implemented by Generative Adversarial Networks (GAN) \citep{goodfellow2014generative}: the search in TSCM space aims to fit a causal model that generates data that decrease discrimination, while the search in the discriminator space (adversary) aims to increase it. In GANs, both spaces are defined by Deep Neural Networks (DNN) and are differentiable; ACT as presented, implements discrete searches and is algorithm-agnostic, not tied to DNNs. The size of the simulated data \(S_C\) is set to the size of the input dataset \(T\) to avoid imbalanced classification problems. 

\begin{figure}
    \centering
    \includegraphics[width=\linewidth]{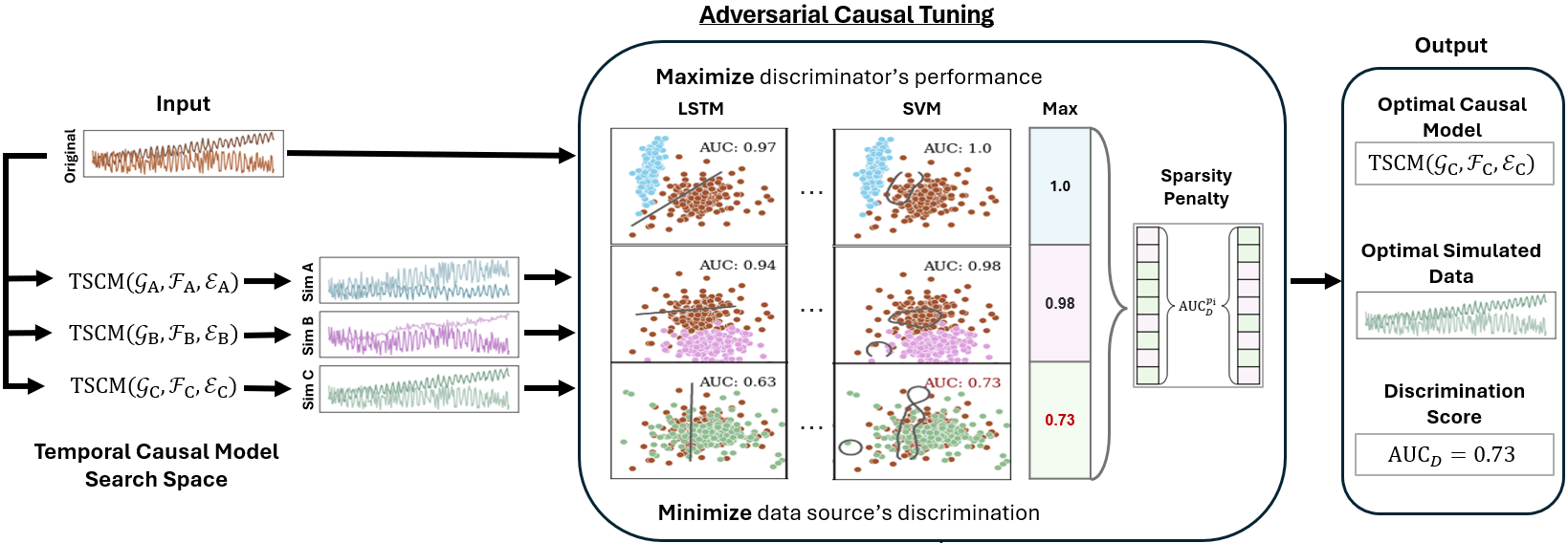}
    \caption{ACT production of discriminator space and selection of the final TSCM: For each TSCM, the discriminator with maximal power is selected. The best performing TSCM is the one with the minimum maximum discrimination performance (TSCM \(C\), third row). The sparsity penalty is applied to select the final model. Its discrimination score is also returned as an estimate of goodness-of-fit.}
    \label{fig:act}
\end{figure}

\paragraph{Sparsity Penalty:} The best performing TSCM \(C_*\) may be overfitting the data, be overly complex, or uninformative. For example, a TSCM with a full causal graph can fit any data distribution. To avoid this situation, it is important to impose a complexity penalty on the selected TSCM. We adopt the same penalty scheme developed for optimizing causal models as in \citet{biza2022oct}. ACT selects as optimal the simplest TSCM, denoted by \(C_{ACT}\), whose discrimination performance is statistically indistinguishable from \(d_{C_*}\). As a simplicity metric for a TSCM \(\langle G, \mathcal{F}, \mathcal{E} \rangle\), we use the number of edges (direct causal relations) in the lagged causal graph, denoted as \(|G|\). To determine statistical equivalence of the performance of a TSCM \(C\) with \(C_*\), we test the null hypothesis \(H:d_{C_*}=d_{C}\) using a permutation-based test \citep{biza2022oct}. Specifically, we consider the vectors of predictions \(v\) and \(v'\) of the best discriminator for \(C_*\) and \(C\) respectively, on the test set. Subsequently, the test returns a p-value \(p_C\) for the hypothesis that the two AUCs \(d_{C_*}\) and \(d_{C}\) are equal. If \(p_C > a\) for some significance level \(a\), the performance of \(C\) is deemed statistically indistinguishable from the best performance achieved. In the implementation, we employ the standard level \(a=0.05\). The TSCM selected by ACT is:
\begin{equation}
\label{eq:act}
    C_{ACT} = \arg_C\min_{C=\langle G, \mathcal{F}, \mathcal{E} \rangle\in\mathcal{C}} |G|, \text { s.t. } p_C > a
\end{equation}

\paragraph{Goodness-of-fit:} ACT returns as a metric of the successful fitting of the input dataset \(T\) with a causal model, the discrimination AUC performance of \(C_{ACT}\) on the test set, namely \(d_{C_{ACT}}\). The closer it is to 0.5, corresponding to random guessing, the better the fit. Values significantly lower than 0.5 indicate good discrimination with the classification labels inverted and should not occur.

\paragraph{Search algorithm:} ACT's uses a simple grid search algorithm over discrete TSCM and discriminator spaces. Figures \ref{fig:overall_architecture_subs} and \ref{fig:act} graphically depict the ACT procedure. Table \ref{tab:search-spaces-0} in appendix \ref{app:search_spaces} summarizes the search spaces. 

\paragraph{Generation of the TSCM Space:} The causal pipelines in \(\mathcal{C}\) include three phases producing a causal graph, functional dependencies, and noise distributions, respectively. In phase I, ACT runs all combinations of available causal discovery algorithms and their hyperparameter values, within prespecified sets of possible values. The current implementation supports methods spanning constraint-based, continuous-optimization, and neural approaches (e.g., PCMCI \citep{runge2018causal}, DYNOTEARS \citep{pamfil2020dynotears}, and CausalPretraining - CP \citep{stein2024embracing}). In phase II, given a lagged causal graph \(G\), functions \(f^j(Pa(V^j_t)\) are fit using regression and time-series forecasting models, such as Random Forests \citep{breiman2001random} and time-series forecasting foundational models \citep{das2024timesfm}. In phase III, for each graph \(G\) and functions \(\mathcal{F}=\{f^j\}\) the residuals \(r^j_i\) are computed for each variable and time-point. The distribution \(\epsilon^j\) of residuals for each \(V^j\) is modeled using density estimators (e.g., flow-based models \citep{rezende2015variational, dinh2016density}) and parametric distributions. At the end of all phases, TSCMs \(\langle G, \mathcal{F}, \mathcal{E} \rangle\) are produced for every combination of algorithm and hyperparameter value for all phases (\textbf{$\approx$700 TSCMs}).

\paragraph{Generation of the Discriminator space \(\mathcal{D}\):} The discriminator space \(\mathcal{D}\) can be produced by any pipeline that accepts real and simulated data \(T\) and \(S_c\), respectively, and outputs a discrimination (binary classification) model. The current implementation supports pipelines with single algorithms, namely Support Vector Machines \citep{hearst1998support} and LSTMs \citep{hochreiter1997long} producing \textbf{\(\approx\)150 discriminators}. 

\section{Experimental Comparative Evaluation of ACT\label{sec:experiments}}

\paragraph{Datasets \label{sec:datasets}:}
We evaluate ACT on corpora of \textbf{real, semi-synthetic, and synthetic} datasets. \textbf{Seven real-world data} are drawn from commonly used benchmarks in the forecasting literature \citep{hahn2023time}, spanning energy, weather, transportation, environmental sensing, and air-quality monitoring. These include the ETTh1/ETTm1 electricity transformer datasets \citep{wu2021autoformer}, the WTH weather dataset \citep{Jiang2023fecam}, AirQualityUCI (AQUCI), hourly bike–pedestrian counts from the Burke--Gilman Trail (BU), outdoor sensor data from Nanning (OD), and the multi-station Air\_Quality dataset used in \citet{cheng2024causaltime} (AQ). Dataset descriptions are provided in Appendix \ref{app:datasets}. We impute missing values through interpolation as in \textit{CausalTime}. Each time-series stationarity has been confirmed by the Augmented Dickey-Fuller \citep{dickey1979distribution} unit root test. To provide standard errors for the computed metrics for each dataset, we sample 10 times 2000 timepoints, similar to Block Bootstrap \citep{gonccalves2011discussion, hardle2003bootstrap}. \textbf{Semi-synthetic systems} are mathematical models (ordinary or partial differential equations) with known causal structure of real physical systems or processes, which are then employed to simulate data. These consist of financial time-series generated from the Fama--French model \citep{huang2015fast, nauta2019causal} (5 datasets), and 6 nonlinear \(f\)MRI BOLD simulations constructed via balloon models \citep{smith2011network, buxton1998dynamics}. Finally, we produce \textbf{20 synthetic causal models} and corresponding datasets by varying the following parameters. A random graph is sampled from either Erd\H{o}s-R\'enyi \citep{brouillard2020differentiable} or Barab\'asi-Albert \citep{barabasi1999emergence} scale-free models. Functional relationships are linear combinations of functions \(g\) of the form \(V^j_t=\sum \alpha_i g(X_i)\), where the coefficients \(\alpha_i\) are randomly sampled, and \(g\) belongs to the set of identity, power, exponential, sigmoid, ReLU, and trigonometric functions, leading to \textbf{non-linear} functional dependencies (except when $g$ is selected to be the identity). Noise distributions are either uniform or Gaussian. The maximum lag \(\ell_\text{max}\in [1,3]\), the number of variables in \([3,12]\), and datasets are simulated for 1000 timepoints (excluding the warmup timepoints). Finally, if a model simulates data with a trend, we resample the functional dependencies until their mean is stationary, similar to \citet{stein2024embracing}. More details on the characteristics of the considered datasets can be found at Appendix \ref{app:datasets}.

\begin{table}[t!]
\small
\centering
\caption{\underline{Comparison of unoptimized MMD vs \(\mathrm{AUC}_D\)}. There are instances (annotated with bold \textcolor{red}{\textbf{red}} color) where \textit{MMD} values are low but \(\mathrm{AUC}_D\) scores indicate high discrimination. Unoptimized MMD, commonly used in the literature, could lead to falsely conclude successful generation of realistic time series data.}
\vspace{10pt}
\setlength{\tabcolsep}{1pt}
\begin{tabular}{lcccccccccc}
\toprule
\multirow{2}{*}{Metric} & \multicolumn{7}{c}{Real} & \multicolumn{2}{c}{Semi-synthetic} & \multirow{2}{*}{Synthetic}\\
\cmidrule(lr){2-8} \cmidrule(lr){9-10}
& WTH 
& AQUCI 
& AQ 
& ETTh1 
& ETTm1 
& BU 
& OD 
& \(f\)MRI 
& Finance 
&  \\
\midrule
MMD 
& \(\textcolor{red}{\mathbf{0.28}} \scriptscriptstyle{\pm .07}\) 
& \(0.46 \scriptscriptstyle{\pm .02}\) 
& \(1.30 \scriptscriptstyle{\pm .74}\)
& \(0.75 \scriptscriptstyle{\pm .16}\) 
& \(\textcolor{red}{\mathbf{0.27}} \scriptscriptstyle{\pm .02}\) 
& \(\textcolor{red}{\mathbf{0.23}} \scriptscriptstyle{\pm .02}\) 
& \(2.36 \scriptscriptstyle{\pm .00}\) 
& \(0.08 \scriptscriptstyle{\pm .01}\) 
& \(\textcolor{red}{\mathbf{0.26}} \scriptscriptstyle{\pm .02}\) 
& \(0.09 \scriptscriptstyle{\pm .11}\) \\
\(\mathrm{AUC}_D\) 
& \(\textcolor{red}{\mathbf{0.85}} \scriptscriptstyle{\pm .01}\)
& \(0.99 \scriptscriptstyle{\pm .00}\)
& \(0.97 \scriptscriptstyle{\pm .00}\)
& \(0.92 \scriptscriptstyle{\pm .02}\)
& \(\textcolor{red}{\mathbf{0.93}} \scriptscriptstyle{\pm .00}\)
& \(\textcolor{red}{\mathbf{0.88}} \scriptscriptstyle{\pm .01}\)
& \(0.99 \scriptscriptstyle{\pm .00}\)
& \(0.71 \scriptscriptstyle{\pm .01}\)
& \(\textcolor{red}{\mathbf{0.99}} \scriptscriptstyle{\pm .00}\)
& \(0.57 \scriptscriptstyle{\pm .00}\) \\
\bottomrule
\end{tabular}
\label{tab:comparison-MMD-AUC}
\end{table}

\paragraph{Metrics \label{sec:metrics}:} 
We use two metrics for comparing distributions: (a) the discrimination performance on the test set, measured by the AUC; in figures and tables, denoted as \(\mathrm{AUC}_D\) to explicitly show that it measures the AUC. (b) The \textit{Maximum Mean Discrepancy -- MMD} \citep{gretton2012kernel} is a metric of distributional discrimination. In both cases, \textit{lower is better} (\(\mathrm{AUC}_D\)'s lowest value is 0.5). For measuring causal graph similarity, we will use (i) the \textit{Structural Hamming Distance (SHD)} \citep{tsam2006maxmin} that counts the number of add or remove an edge or reverse orientation operations to transform the PDAG of the estimated graph to the PDAG of the true graph (\textit{lower is better}); (ii) \textit{Adjacency AUC} (\(\mathrm{AUC}_{\text{adj}}\)), following \citet{cheng2024causaltime, stein2024embracing}: we rank edges in a graph according to a causal model's confidence and compute the AUC. Higher values are better; they indicate that the causal model about the edges present in the true graph. 

\paragraph{Unoptimized MMD Can be a Misleading Metric of Discrimination:} Table \ref{tab:comparison-MMD-AUC} shows the unoptimized MMD metric using the default parameters (Gaussian kernel) against the \(\mathrm{AUC}_D\) for the corpus datasets, including 10 synthetic ones. Red fonts indicate situations where MMD is relatively low (\(<0.3\)) but \(\mathrm{AUC}_D\) significantly higher than \(0.5\). Standard deviation on the real datasets has been produced by repeating the experiment \(10\) times. Such low MMDs (\(0.21\) to \(0.24\)) led \citet{cheng2024causaltime} to conclude that CausalTime has been validated "to generate realistic timeseries".

\paragraph{Evaluation of the Estimated Causal Structure:} \label{subsubsec:sparsity_experiment}

\begin{figure*}
    \centering
    \begin{minipage}[t]{1.0\textwidth}
        \centering
        \includegraphics[width=\textwidth]{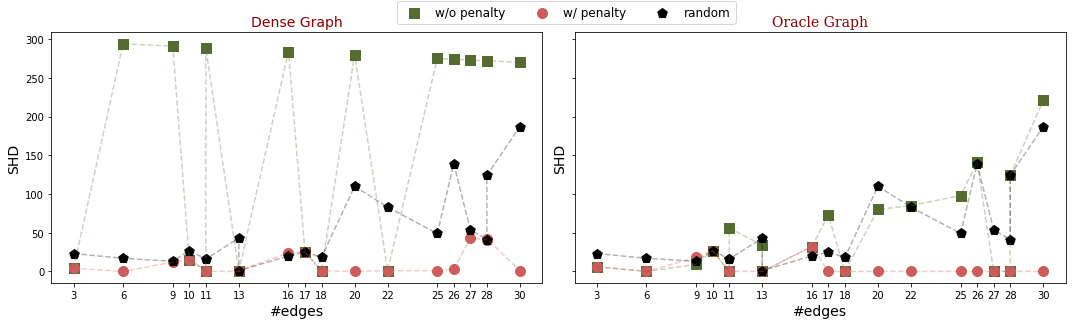}
    \end{minipage}%
    \hfill
    \caption{SHD of ACT with/without penalty and random selection of causal model on 19 synthetic datasets. \textbf{(a)} TSCM search space includes the fully connected graph. ACT with sparsity penalty avoids selecting the dense model in all cases. \textbf{(b)} TSCM search space includes the true causal graph. ACT with sparsity penalty selects it in 16/19 cases, compared to 7/19 without the penalty. ACT's selections are always better than random guessing. }
    \label{fig:sparsity_penalty}
\end{figure*}

We evaluate ACT as a causal model selection method. We compare the SHD of ACT without the sparsity penalty (i.e., selecting the best performing model \(C_*\), Eq. \ref{eq:best-performing})), with the penalty (\(C_{ACT}\), Eq. \ref{eq:act}), and without using ACT at all, but choosing a TSCM at random out of all TSCMs considered. Results are shown on 19 synthetic datasets with 10 variables and a varying number of edges in their lagged graph. In Figure \ref{fig:sparsity_penalty}(Left) we show the SHD when we add to the TSCM search space \(\mathcal{C}\) the fully connected graph (Dense Graph). Without the sparsity penalty, ACT overfits the graph selection (high SHD, green line) selecting the full graph 8 times, while ACT with the penalty avoids it in all cases. In Figure \ref{fig:sparsity_penalty}(Right) we add to the TSCM search space the true causal graph. We observe that ACT with the sparsity penalty selects it 16/19 cases, compared to 7/19 cases without the penalty. Overall, we conclude that \textit{ACT (i) consistently avoids the trivial fully connected solution, (ii) selects the true graph causal graph if it is included in its search space, (iii) is always better than selecting a TSCM randomly within its search space}. 

\paragraph{Comparison to Temporal Data Generation Methods for Observational Distributions} \label{subsubsec:comparison}
We compare the efficacy of the methodologies in Section \ref{sec:related-work} against ACT in generating realistic datasets. At present, we consider datasets generated from the observational distribution, i.e., without simulating the effect of interventions. We compare against the authors' implementation of \textit{CausalTime}, \textit{CPAR} as implemented by \textit{SDV} \citep{patki2016synthetic}, and TimeVAE \citep{desai2021timevae} available through the Synthcity \citep{qian2023synthcity} framework. Figure \ref{fig:act-vs-sota-generation} shows the results. We observe that \textit{ACT outperforms all other methodologies on the synthetic datasets, even though it attempts to solve a harder problem, i.e., fitting a generative causal model instead of a non-causal generative model}. On the corpus of real datasets, all methodologies fail to generate realistic datasets, with CausalTime obtaining better performance. Detailed results on the semi-synthetic datasets are omited for brevity; all methodologies are on par on the \(f\)MRI datasets (\(\mathrm{AUC}_D\) in [0.71, 0.78]) and fail on the finance datasets (\(\mathrm{AUC}_D\) in [0.78, 0.99]), with ACT and CausalTime being the winners in each corpus of semi-synthetic datasets, respectively. We note again, that despite its name, CausalTime does not produce a lagged causal graph, but a summary graph, thus solving an easier problem. 

\begin{figure}[t!]
    \centering
    \begin{minipage}[t]{0.42\textwidth}
        \centering
        \includegraphics[width=\textwidth]{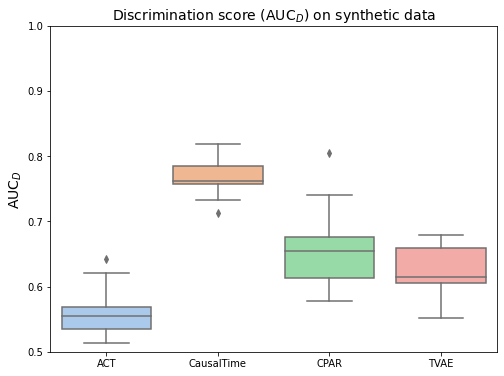}
    \end{minipage}%
    \hfill
    \begin{minipage}[t]{0.42\textwidth}
        \centering
        \includegraphics[width=\textwidth]{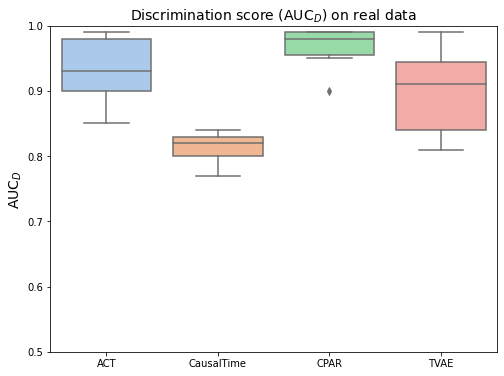}
    \end{minipage}
    \caption{Discrimination performances of state-of-the-art on the corpora or synthetic (left) and real (right) datasets. ACT outperforms all benchmark methods on the synthetic datasets. All methods fail to capture the input data distribution on the real data. ACT is the only methodology that captures the data distribution by fitting a causal model, thus solving a harder problem.}
    \label{fig:act-vs-sota-generation}
\end{figure}

\paragraph{ML Efficacy \label{subsubsec:ml-efficacy}:} We evaluate the ML Efficacy of ACT and CausalTime. ML efficacy is measured with the \(\mathrm{AUC}_{adj}\), comparing the ranking (according to confidence) of discovered causal edges using the generated data against the one using the original data as the ground truth. A high ML efficacy score implies that one could share the generated data instead of the real ones, and still obtain similar results in terms of causal discovery. To learn the causal models, we use PCMCI with the default parameters. The average \(\mathrm{AUC}_{adj}\) on the real, semi-synthetic, and synthetic corpora is 0.66, 0.74, and 0.98 for ACT and 0.60, 0.72, and 0.88 for CausalTime, respectively. Overall, results indicate that when fitting a causal model is successful (low \(\mathrm{AUC}_{D}\)), such as in the case of synthetic data, ML efficacy is achieved (high \(\mathrm{AUC}_{adj}\)). Detailed results can be seen at appendix \ref{app:ml_effic}. 

\paragraph{Temporal Data Generation for Interventional Distributions:} \label{subsubsec:interventions} A key advantage of fitting a causal model is the ability to generate data under interventional conditions. We evaluate realistic interventional data generation on 29 synthetic causal models with 5-10 variables and up to 3 lags. ACT is applied to learn an optimal causal model from 1000 observational time-points. Interventional datasets of 1000 time-points are generated by using the true causal model and the estimated one by ACT. To simulate interventions, a time point within [1, 750] is selected, and an intervention is simulated by setting the value of one of the variables to its minimum or maximum in the observational data set. Hence, we simulate a hard intervention at a single time point that resembles a fault-injection. Figure \ref{fig:interventions} shows the optimized observational and interventional discrimination AUCs. In \(19/29\) cases, the observational and interventional \(\mathrm{AUC}_{D}\)s differ by less than \(0.1\), indicating that interventional data under the estimated causal model resemble the corresponding intervention by the true model. In \(4/29\) the \(\mathrm{AUC}_{D}\) difference is in \([0.1, 0.2]\), and in \(6/29\) larger than \(0.2\). The results indicate that in most cases, the TSCM learned by ACT could serve as a causal digital twin that correctly predicts the effect of experiments. We have explored the cases where the interventional discrimination is large and noticed that it could happen even when the causal structure is learned perfectly from the observational data. One explanation for the occasional large differences is that they unavoidable; specifically, interventional datasets may have different distributions even for the same model and intervention, depending on the initial conditions due to non-linear dynamics. We intend to further explore this issue. Finally, notice that a comparison with the other generative methodologies is impossible since none of them can simulate interventions; non-causal models could potentially simulate conditional distributions, but that would require non-trivial adaptations of related work.

\begin{figure} 
    \centering
    \includegraphics[width=0.70\textwidth]{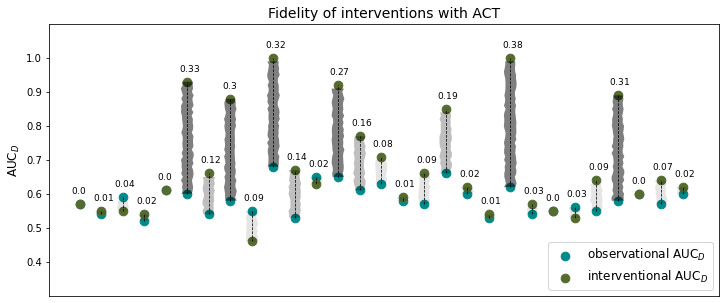}
    \caption{Comparison between the optimized \(\mathrm{AUC}_{D}\) of the model selected by ACT    
    on observational and interventional datasets, for 29 synthetic TSCMs. The vertical lines are labeled with the \(\mathrm{AUC}_{D}\) differences. In 19/29 experiments, the differences are smaller than 0.1 (light gray shade). This indicates the TSCM learned from observational data, generates interventional data similar to the data produced by the true model.}

    \label{fig:interventions}
\end{figure}

\section{Conclusions, Limitations, and Discussion}
\label{sec:conclusion}

We introduce Adversarial Causal Tuning (ACT), which can be used for timeseries generation under either observational or interventional settings, creation of benchmark causal models, and causal model selection. Experimental results on real, synthetic, and semi-synthetic datasets show that penalizing for complexity is paramount in causal model selection. ACT outperforms the state-of-the-art of non-causal time series data generation on synthetic models, despite solving a harder problem. However, all proposed algorithms fail to successfully fit distributions of real time series data; this still remains an open research problem. ACT returns an optimized metric of goodness-of-fit that warns regarding unsuccessful fitting. In contrast, some prior works do not optimize the discrimination metric and are led to falsely conclude that successful fitting has been achieved. 

ACT is currently limited to time series data under the assumptions of additive noise, causal sufficiency, and no contemporaneous causalities. To improve performance on real datasets, it may be necessary to remove these assumptions. This is not trivial: for example, fitting functional relationships in the presence of latent confounding variables is challenging. ACT can benefit from including better causal discovery algorithms in its search space, as it is shown that when it is presented with the true causal graph, it manages to successfully identify it among dozens of other alternatives. It can also benefit from employing better search techniques as Bayesian Optimization and meta-level learning adapted for minimax problems. Other improvements are possible: estimation of discrimination performance can be improved by using (repeated) cross-validation for small sample sizes. One could generate multiple simulated datasets \(S_C\) and take the average discrimination over all of them for better estimation, sacrificing computational efficiency. We note that the discrimination performance reported stems from a minimax procedure, which may introduce estimation bias. Techniques such as Bootstrap Bias Correction (BBC) \citep{tsamardinos2018bootstrapping} should be adopted to correct the returned AUCs for this bias. Discrimination pipelines could be extended by including other types of algorithms than classifiers, such as time series selection \citep{vareille2024chronoepilogi}, providing additional information regarding the variables that create the distributional differences.

\section{Reproducibility Statement \label{sec:reproducibility}}

The complete source code, including documentation, implementation, and benchmarking scripts, is publicly available at \href{}{[removed for anonymous submission]}.


\acks{We would publicly like to thank Konstantina Biza and \'Etienne Vareille for their fruitful comments throughout this work.}

\bibliography{references}

\newpage

\appendix

\section{Search Spaces\label{app:search_spaces}}

\begin{table*}[h!]
\centering
\caption{Search spaces for Temporal Causal Modeling (TSCM, left) and Discriminator (C2ST, right). Overall approximately \textbf{700 TSCMs} and \textbf{150} discriminators are searched over.}
\label{tab:search-spaces-0}

\renewcommand{\arraystretch}{1.05}

\begin{tabular*}{\textwidth}{@{\extracolsep{\fill}} p{0.60\textwidth} p{0.38\textwidth} }
\textbf{(a) TSCM Search Space} & \textbf{(b) Discriminator Search Space} \\
\midrule[1pt]

{\small
\setlength{\tabcolsep}{3pt}
\begin{tabular}{@{}llll@{}}
\textbf{Phase} &\textbf{Method} & \textbf{Param.} & \textbf{Values} \\
\midrule[0.66pt]
& CP & Model size & \texttt{398M} (2 models) \\
Phase 1 & PCMCI & \(\ell_{\text{max}}\)/iters & 2, 3 / 10 \\
& DYNOTEARS & \(\ell_{\text{max}}\) & 1, 3 \\
\midrule[0.33pt]
& RF & \(n_\text{estimators}\) & 100, 500, 1000 \\
Phase 2 & TCDF & Levels / ks / \(lr\) & 0, 2 / 2, 3 / 1e–2, 1e–3 \\
& TimesFM & Model & \texttt{timesfm-1.0-200m} \\
\midrule[0.33pt]
& Gaussian & \(\mu\) / \(\sigma\) & empirical \\
& Uniform  & \(a\) / \(b\) & empirical \\
Phase 3 & Residuals & - & empirical \\
& Splines  & \(lr\) / bins & 1e-2 / 16 \\
& RealNVP  & \(lr\) & 1e-4 \\
\end{tabular}
}
&
{\small
\setlength{\tabcolsep}{3pt}
\begin{tabular}{@{}lll@{}}
\textbf{Method} & \textbf{Param.} & \textbf{Values} \\
\midrule[0.66pt]
SVM & C & 1.0, 0.75, 0.5, 0.25 \\
 & Kernel & linear, poly, rbf \\
 & Degree & 3 \\
 & Gamma & auto, scale \\
\midrule[0.33pt]
LSTM & Batch size & 32, 64 \\
 & Hidden dim & 128, 256 \\
 & Layers & 2, 3 \\
 & Dropout & 0.05, 0.1 \\
 & SeqLen & 10, 20 \\
 & Epochs & 10, 50 \\
 & LR & 1e–4, 1e–3 \\
\end{tabular}
}
\end{tabular*}

\end{table*}

\section{Datasets\label{app:datasets}}

\begin{table}[h!]
\centering
\caption{Overview of Real Time-Series Datasets.}
\vspace{10pt}
\begin{tabular}{|c|c|c|c|c|c|}
\hline
\textbf{Dataset} & \textbf{Variables} & \textbf{Timesteps} & \textbf{Granularity} & \textbf{Start Date} & \textbf{Domain} \\ 
\hline
ETTh1 & 7 & 17420 & 1 hour & 07/01/2016 & Power \\ 
ETTm1 & 7 & 69680 & 15 min & 07/01/2016 & Power \\ 
WTH & 12 & 35064 & 1 hour & 01/01/2020 & Weather \\ 
Air\_Quality & 36 & 8760 & 1 hour & - & Weather \\
AirQualityUCI & 12 & 9357 & 1 hour & 03/10/2024 & Weather \\
Bike-usage & 5 & 552584 & 1 hour & 01/01/2014 & Transportation \\
Outdoors & 3 & 1440 & 1 sec & 21/02/2016 & Environmental \\
\hline
\end{tabular}
\label{tab:real-datasets}
\end{table}

\begin{table}[h!]
\centering
\caption{Overview of the Synthetic Time-Series Dataset Collection.}
\vspace{10pt}
\begin{tabular}{|c|c|c|c|c|c|}
\hline
\textbf{Collection} & \textbf{Num. of Datasets} & \textbf{Variables} & \textbf{Timesteps} & \textbf{Max Lag} & \textbf{Functional Relationships} \\ 
\hline
C1 & 10 & 3-12 & 1000 & 1-3 & Linear \& Non-linear \\ 
\hline
\end{tabular}
\label{tab:synthetic-datasets}
\end{table}

\begin{table}[h!]
\centering
\caption{Overview of Semi-synthetic Time-Series Dataset Collections.}
\vspace{10pt}
\begin{tabular}{|c|c|c|c|c|c|}
\hline
\textbf{Collection} & \textbf{Num. of Datasets} & \textbf{Variables} & \textbf{Timesteps} & \textbf{Max Lag} & \textbf{Functional Relationships} \\  
\hline
\(f\)MRI & 6 & 5 & 1200 - 5000 & 1 & Non-linear \\ 
Finance & 5 & 25 & 4000 & 1 & Non-linear \\ 
\hline
\end{tabular}
\label{tab:semi-synthetic-datasets}
\end{table}

\section{ML Efficacy\label{app:ml_effic}}

\begin{table}[h!]
\centering
\caption{\underline{ML Efficacy:} \(\mathrm{AUC}_{\text{adj}}\) of the causal graphs derived from simulated data against the ground truth. A high \(\mathrm{AUC}_{\text{adj}}\) score indicates the simulated data act similar to the original on CD tasks. In most cases \textit{ACT} clearly outperforms \textit{CausalTime} (see \textit{ETTh1}, \textit{ETTm1}, \textit{BU}, \textit{OD}, \textit{fMRI} and \textit{Synthetic}). There are 3 cases where \textit{CausalTime} scores higher than \textit{ACT} (\textit{WTH}, \textit{AQUCI} and \textit{Finance}), but the difference is to the second decimal. Another observation is that both methodologies seem incapable of capturing the behavior of the original data for certain cases (\textit{WTH}, \textit{AQ}, \textit{OD} and \textit{Finance}), a realization that aligns with our remarks about the strain of current the current state-of-the-art to causally simulate real time-series.}
\label{tab:ml-efficacy}
\setlength{\tabcolsep}{2pt}
\begin{tabular}{lcccccccccc}
\toprule
\multirow{2}{*}{Method} & \multicolumn{7}{c}{Real} & \multicolumn{2}{c}{Semi-synthetic} & \multirow{2}{*}{Synthetic}\\
\cmidrule(lr){2-8} \cmidrule(lr){9-10}
& WTH 
& AQUCI 
& AQ 
& ETTh1 
& ETTm1 
& BU 
& OD 
& \(f\)MRI 
& Finance 
&  \\
\midrule

ACT
& \(0.58\)
& \(0.72\)
& \(0.56\)
& \textcolor{red}{\(\textbf{0.76}\)}
& \textcolor{red}{\(\textbf{0.77}\)}
& \textcolor{red}{\(\textbf{0.72}\)}
& \textcolor{red}{\(\textbf{0.50}\)}
& \textcolor{red}{\(\textbf{0.86}\)}
& \(0.61\)
& \textcolor{red}{\(\textbf{0.98}\)} \\

CausalTime
& \textcolor{red}{\(\textbf{0.59}\)}
& \textcolor{red}{\(\textbf{0.73}\)}
& \(0.56\)
& \(0.59\)
& \(0.68\)
& \(0.58\)
& \(0.48\)
& \(0.77\)
& \textcolor{red}{\(\textbf{0.67}\)}
& \(0.88\) \\

\bottomrule
\end{tabular}

\label{tab:ml-efficacy-transposed}
\end{table}

\end{document}